\documentclass[runningheads]{llncs}

\usepackage{eccv}

\usepackage{eccvabbrv}
\usepackage{microtype}
\usepackage{graphicx}
\usepackage{subcaption}
\usepackage{booktabs}
\usepackage{placeins}
\usepackage[pagebackref,breaklinks,colorlinks,citecolor=eccvblue]{hyperref}
\usepackage{orcidlink}

\newcommand{\method}{StateKV\xspace}
\newcommand{\Method}{StateKV\xspace}
\newcommand{\sliding}{ReKV\xspace}

\usepackage{amsmath}
\usepackage{amssymb}
\usepackage{mathtools}

\usepackage{amsthm}

\usepackage[capitalize,noabbrev]{cleveref}

\makeatletter
\@ifclassloaded{llncs}{
}{
  \theoremstyle{plain}

  \theoremstyle{definition}

  \theoremstyle{remark}
  
}
\makeatother

\usepackage[textsize=tiny]{todonotes}
\usepackage{xspace}
\usepackage{xparse}
\usepackage{multirow}
\usepackage{pgfplots}
\pgfplotsset{compat=1.18}

\NewDocumentCommand{\slurm}{m g}{%
  \IfNoValueTF{#2}{\texttt{S:#1}}{\pgfmathprintnumber[fixed,precision=2]{#2}\%}%
}

\begin{document}

\title{Linear Scaling Video VLMs for Long Video Understanding}
\titlerunning{Linear Scaling Video VLMs for Long Video Understanding}

\author{
Crist\'obal Eyzaguirre\inst{1} \and
Jiajun Wu\inst{1} \and
Juan Carlos Niebles\inst{1}
}
\institute{Stanford University\\  \url{https://ceyzaguirre4.github.io/StateKV}}


\authorrunning{C. Eyzaguirre et al.}
\maketitle

\begin{abstract}
  Video vision-language models (VLMs) are increasingly used in long-horizon and streaming settings, yet most video encoders still rely on spatiotemporal self-attention, causing compute and latency to grow quadratically with the number of frames.
  Existing efficiency methods improve scalability but often lose accuracy relative to full self-attention, for example through aggressive frame/token dropping or coarse attention approximations.
  We introduce \method, an inference-time method that adapts pretrained long-video VLMs to linear-time video prefill by carrying cross-frame context in a fixed-capacity, importance-based recurrent state,  paired with a second full per-frame cache used for decoding.
  Across three long-video benchmarks and seven models spanning three families and multiple scales, \method remains close to full self-attention and consistently outperforms dominant sliding-window / recency-based streaming approximations, without fine-tuning or architectural changes.
  \Method also reduces video-prefill cost measured FLOPs, enabling stronger accuracy at a fixed compute budget by running larger models.
  These results suggest a practical step toward scalable long-video understanding.
\end{abstract}

\begin{figure}[ht]
  \vskip -0.4in
  \begin{center}
    \centerline{\includegraphics[page=1,width=\columnwidth]{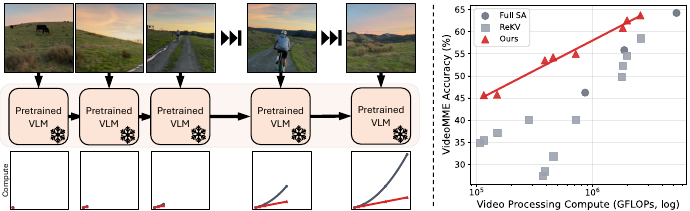}}
    \caption{Overview of \method. Left: a frozen pretrained VLM processes a video incrementally, one frame at a time. While \method maintains approximately constant marginal video-prefill compute per added frame (red), unmodified full self-attention (gray) incurs in increasing per-frame cost. This yields linear video-prefill scaling in the number of frames for \method, in contrast to quadratic scaling for the base model. Right:  compute-accuracy frontier at 512 frames (VideoMME), where \method surpasses \sliding and full self-attention operating points across practical compute budgets.}
    \vskip -0.4in
    \label{fig:overview}
  \end{center}
\end{figure}

\section{Introduction}

Modern video VLMs are increasingly vital for long-horizon and streaming tasks such as autonomous driving and embodied robotics, where models must integrate evidence over minutes or hours, often in real time.
However, a central challenge for long-video VLMs  is that their computational cost grows quadratically with the length of the video.
Because dominant architectures allow each frame to attend over all previous video tokens, the per-frame computational cost rises as the video grows, and the overall complexity of processing the entire video scales quadratically.
This creates a significant bottleneck for practical deployment and rules out real-time streaming applications entirely: a car that has driven for an hour is, in principle, harder to query than one that just started.
Linear overall complexity or, equivalently, constant per-frame cost, is therefore central to scalable long-video understanding, and a critical prerequisite for streaming.

Most existing efficiency strategies focus on shrinking the input to the underlying quadratically-scaling transformer, but this often reduces information more than it reduces true complexity.
Common approaches (i) subsample frames, (ii) trim visual tokens at the input layer, or (iii) compress the visual context (e.g. by compressing the KV cache).
A practical limitation of these methods is that aggressive compression can substantially degrade long-video performance unless a relatively large fraction of the visual information is retained.
For instance, prior work found they must keep a large token fraction (around 60\%) to avoid severe degradation when using cache-compression~\cite{wang2025adaretake}.
This issue can be even more acute for frame dropping or aggressive patch trimming, which may remove temporal and spatial cues required for long-horizon reasoning.
As a result, many frame/patch/token dropping methods improve efficiency mainly by redistributing quadratic cost over a shorter sequence rather than changing the scaling law.

Streaming-prefill methods offer a stronger tradeoff by avoiding compression altogether, instead separating video encoding from text generation and using heuristics to reduce the computational overhead of  generating the video-only KV cache.
Methods such as \sliding adapt pretrained VLMs, and rely on sliding-window-based mechanisms to process frames sequentially while constructing the video KV cache~\cite{rekv2025}.
Unlike fixed-budget compression methods whose final generation context is $O(1)$ in the number of frames, these approaches usually keep all per-frame visual tokens for decoding, so generation remains $O(N)$ in the number of frames $N$.
Although this means generation is less efficient, this tradeoff is often favorable in long-video settings because the dominant cost by far is the video-prefill stage, and its complexity is reduced from $O(N^2)$ to $O(N)$, yielding end-to-end complexity that is linear in the number of frames.
Furthermore, follow-up work has explored how to reduce the latency of the \emph{generation/query} stage on top of this streaming-prefill paradigm, making it suitable for real time applications~\cite{hermes2026,livevlm2025,streamkv2025}.

While effective in many cases, recency-based heuristics are largely ad-hoc.
Our contribution comes from framing streaming video prefill on frozen pretrained backbones as approximating full self-attention with a small set of tokens carried between frames.
This formulation makes weaker modeling assumptions than a strict recency window and leads to a more principled question: which information must be preserved between frames so that streaming prefill remains a good approximation to full attention?
Our key observation is that long-video attention in pretrained VLMs  is highly structured: most interactions are within-frame, while long-range temporal interactions often concentrate on a small set of ``temporal sink'' tokens that evolve slowly over time. This structure appears in the tested cases we analyze, is consistent with how VLM backbones are usually trained (image-first, then adapted to videos with variable numbers of frames), and aligns with prior work documenting attention-sink phenomena in language transformers~\cite{xiao2023streamingllm,gu2024attentionsink}, on which VLMs are built.
This motivates our novel approach: retain all per-frame tokens for final decoding as in prior work, but reduce the complexity of video-prefill by limiting long-range cross-frame interactions with a fixed-capacity state that identifies and preserves locally relevant tokens and temporal sinks. 

We introduce \method, an inference-time KV-cache prefill method that adapts a frozen pretrained VLM backbone to this self-attention-approximation view of streaming video prefill.
Its design follows a small set of core assumptions 
and is implemented via two coupled caches per layer: a fixed-capacity temporal state for cross-frame context, and a detailed per-frame cache that preserves intraframe structure.
\Method builds the video KV cache incrementally as frames arrive, then performs standard text decoding conditioned on all video tokens.
This yields $O(N)$ video encoding while preserving full per-frame detail for decoding, resulting in end-to-end VideoQA complexity that is linear in the number of frames.

Empirically, \method delivers strong long-context results on three long-video benchmarks and consistently outperforms the dominant sliding-window / recency-based streaming approximation family used for linear-time video prefill.
Across settings, it more closely matches full ($O(N^2)$) spatiotemporal attention while requiring no fine-tuning or architectural changes.
Across three model families and spanning multiple parameter scales, the same trends recur: \method remains close to full attention and improves steadily as state capacity increases. Moreover, the reduction in measured FLOPs is dramatic enough to change what is achievable at a given compute budget: \method enables running larger, more accurate models for cheaper than full self-attention smaller ones.

\section{Related Work}

\paragraph{Vision language models and the long-video bottleneck.}
Modern vision-language models (VLMs) typically pair a strong image encoder (often CLIP-like~\cite{radford2021clip}) with a large language model through a lightweight cross-modal interface, enabling open-ended visual understanding and instruction following~\cite{videounderstandingsurvey,largescalemultimodalsurvey,alayrac2022flamingo,li2023blip2,liu2023llava,liu2023improvedllava,liu2024llavanext}.
Recent open and open-weight systems such as LongVA~\cite{zhang2024longva}, InternLM-XComposer~\cite{internlmxcomposer,internlmxcomposer2_5}, TimeMarker~\cite{timemarker}, InternVL2.5/3~\cite{chen2024internvl2.5,internvl3_2025}, Qwen-VL/Qwen2-VL/Qwen2.5-VL/Qwen3-VL~\cite{Qwen-VL,Qwen2-VL,Qwen2.5-VL,qwen3vl2025}, Molmo~\cite{li2024molmo}, Eagle2.5~\cite{eagle}, Apollo~\cite{apollo}, and LLaVA-style models including LLaVA-OneVision and LLaVA-Video~\cite{li2024llavaonevision,zhang2024llavavideo} further push general-purpose multimodal reasoning and support multi-image and video inputs.
Extending these models to video requires aggregating information across many frames; representative video-LLMs include Video-LLaMA~\cite{damonlpsg2023videollama}, VideoChat~\cite{2023videochat}, VideoLLaMA3~\cite{damonlpsg2025videollama3}, Video-ChatGPT~\cite{Maaz2023VideoChatGPT}, LITA~\cite{huang2024lita}, Momentor~\cite{qian2024momentor}, HawkEye~\cite{wang2024hawkeye}, and TimeChat~\cite{timechat}, which adapt image-first backbones using temporal pooling, per-frame tokenization, and explicit temporal reasoning modules.
Closed models such as Gemini, GPT-family systems, and Claude demonstrate increasingly strong fine-grained and long-context video understanding~\cite{reid2024gemini,comanici2025gemini,gpt5,Achiam2023GPT4TR,anthropic2025sonnet}.
As video duration grows, the video-side prefill cost  becomes the dominant constraint, motivating methods that either shrink the number of visual tokens presented to the model or change how long-range video context is processed.

\paragraph{Token reduction without changing asymptotic order.}
One line of work improves efficiency by reducing the number of visual tokens per frame or the number of frames presented to the model, while leaving the overall sequence-processing pattern unchanged.
Early evidence for this family came from ATP, which studied how far single-frame selection can go in VideoQA and highlighted that many benchmarks admit surprisingly strong atemporal baselines~\cite{buch2022atp}; more recent analysis in Codeplexity further argues that current VideoQA models struggle most on questions whose latent programs require integrating evidence across multiple frames~\cite{eyzaguirre2025codeplexity}.
At the frame level, methods based on subsampling, adaptive frame selection, or temporal search retain only a subset of frames for downstream reasoning~\cite{buch2022atp,buch2025flexible,tstar2025,yu2023self}.
At the token level, prior work reduces the number of visual tokens through pooling, similarity-based merging, or resampling into a fixed set of latent queries~\cite{xu2024pllava,li2024llamavid,slowfastllava,jin2023chatunivi,li2024videochat,llavamini}, others instead retain a subset of the original tokens, scored by attention or importance (sometimes query-aware)~\cite{fastv,fu2024framefusion,Xing2024PyramidDropAY,zhang2024sparsevlm,flexselect}, while others combine these operations across both space and time~\cite{dycoke,llavascissor}.
Recent surveys organize these methods as part of a broader multimodal token-compression literature~\cite{shao2025tokens}.
Recent codec-aware tokenization approaches such as CoPE-VideoLM~\cite{copevideolm2026} also fit naturally in this family: rather than changing the temporal processing order, they reduce the number of tokens contributed by most frames by replacing dense RGB encoding with compact codec-derived representations.
These approaches often provide strong practical savings, but they primarily shrink the per-frame representation; they do not directly change the growth of cross-frame attention once many frames remain in context.

\paragraph{Long-video and streaming video inference with changed complexity.}
In long videos, thousands of frames with hundreds of visual tokens per frame can yield effective sequence lengths in the millions, making both quadratic attention and the linear-in-length KV cache footprint dominant constraints.
Hybrid long-video architectures such as VAMBA~\cite{ren2025vamba} replace expensive video-token self-attention with linear-time state-space modules to enable long video understanding, but they require architectural changes and costly training.
Also requiring training, StreamingVLM~\cite{xu2025streamingvlm} induces a fixed attention pattern that relies on sink anchors so real-time inference can enforce the same pattern efficiently.

Query-agnostic fixed-budget schemes instead compress or regulate the video-side KV cache to support longer videos or unbounded streams, e.g., InfiniPot-V~\cite{infinipotv2025}, StreamMem~\cite{streammem2025}, and MovieChat~\cite{song2023moviechat,song2024moviechat+}.
In parallel, a recent line of work reduces long-video cost by decoupling video processing (``prefill'') from decoding and constructing a video KV cache incrementally as frames arrive.
ReKV~\cite{rekv2025} is an inference-time adaptation of pretrained VLMs that reduces the complexity of long video encoding via a sliding-window mechanism, then answers questions by decoding conditioned on the accumulated per-frame tokens.
Follow-up methods such as HERMES~\cite{hermes2026}, LiveVLM~\cite{livevlm2025}, and StreamKV~\cite{streamkv2025} further optimize the language generation stage via hierarchical KV memories, streaming-oriented KV cache construction and retrieval, or segment-level retrieval/compression with summary tokens, respectively.
These directions blur the boundary between ``long-video'' and ``streaming'' settings: in both cases, the dominant computation is often the video-side prefill, and improving real-time prefill throughput directly benefits online interaction.
Related streaming systems such as SDQES~\cite{song2024sdqes}, streaming dense video captioning~\cite{zhou2024streamingdensecaption}, VideoLLM-online~\cite{chen2024videollmonline}, Flash-VStream~\cite{zhang2024flashvstream}, and Dispider~\cite{qian2025dispider} further emphasize causal processing and online memory management.

Our approach sits between fixed-budget compression and streaming approximations: we compress the running state used to build the prefill memory, while preserving per-frame visual detail for final language generation.
Unlike \sliding~\cite{rekv2025}, which imposes a strict sliding-window view of the past, we treat streaming video prefill as approximating full self-attention with a small set of tokens carried between frames. This yields a two-cache structure: a fixed-capacity, importance-based temporal cache used only during streaming video prefill that carries the important tokens, and a full detailed cache retained for final language generation.
This design is motivated by empirical structure in long-video attention (dominant within-frame interactions plus a small set of slowly-varying ``temporal sink'' tokens), and we find that it transfers consistently across multiple pretrained model families and parameter scales.

\paragraph{KV-cache compression for long-context LLMs and multimodal models.}
A large literature studies reducing memory and bandwidth costs of the KV cache in long-context LLM inference.
Training-free eviction and sparsification policies such as H$_2$O~\cite{zhang2023h2o} retain a mixture of recent and ``heavy hitter'' tokens, while learned or lightly-trained approaches compress or sparsify the cache online (e.g., DMC~\cite{nawrot2024dmc} and DMS~\cite{lancucki2025dms}).
Other methods design layer- or task-adaptive cache budgets (e.g., PyramidKV~\cite{cai2024pyramidkv}) or provide practical drop-in schemes for decoding acceleration (e.g., SnapKV~\cite{li2024snapkv} and RocketKV~\cite{behnam2025rocketkv}).
For \emph{multimodal} long-context inference, AdaReTaKe~\cite{wang2025adaretake}, Look-M~\cite{wan2024lookm} and MEDA~\cite{wan2025meda} study KV allocation/retention across modalities.
Compared to these mostly text-centric policies, long-video inference places much of the cost in the \emph{video prefill} phase and exhibits distinct attention structure due to frame-based inputs, motivating video-specific designs.

\section{Method}

\begin{figure}[ht]
  \begin{center}
    \centerline{\includegraphics[page=1,width=\columnwidth]{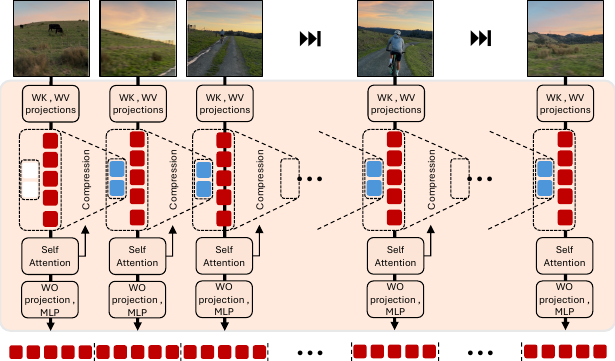}}
    \caption{Single-transformer-layer view of \method, showing the required modifications to a transformer block. The video stream is processed frame-by-frame with
    a frozen backbone. A fixed-capacity compressed state (in blue) allows information from previous frames to flow through a fixed size set of sink tokens during prefill. Separately, we build a full length detailed state for decoding (shown in red).}
    \label{fig:method_diagram}
  \end{center}
\end{figure}

\subsection{Core Assumptions}
\label{sec:assumptions}

Our design relies on empirical properties of attention in long-video VLMs.
Let frame $n$ provide $T$ query tokens and let $\mathcal{H}_{n-1}$ denote the set of keys from frames $< n$. For a fixed layer, let $A_n$ denote the attention weights produced when processing frame $n$. We define the cross-frame importance of a key $j \in \mathcal{H}_{n-1}$ as
{\small\begin{equation}
  s_{n,j} = \sum_{i=1}^{T} A_{n,i,j}.
\end{equation}
}Our first assumption is that there exists a set of temporal sinks $S_n \subseteq \mathcal{H}_{n-1}$ with $|S_n| = K$ and $K \ll |\mathcal{H}_{n-1}|$ such that
{\small\begin{equation}
  \sum_{j \in S_n} s_{n,j}
  \approx
  \sum_{j \in \mathcal{H}_{n-1}} s_{n,j}.
\end{equation}
}That is, most \emph{inter-frame} attention mass is concentrated on a small set of historical tokens whose size does not grow with the total video length. This is the basis for limiting cross-frame attention to a fixed-capacity temporal memory.

Second, we assume that these sink sets evolve slowly over time, in the sense that the next useful sink set can be well approximated from the previous one together with the current frame:
{\small\begin{equation}
  S_{n+1} \approx \mathrm{TopK}\bigl(S_n \cup \{\text{tokens in frame } n+1\}\bigr).
\end{equation}
}This assumption justifies an incremental update rule that evicts low-importance entries and admits newly salient tokens from the current frame, rather than re-optimizing over the entire video prefix at every step.

Finally, because \method retains all video tokens for text decoding, the compressed state only needs to approximate frame-to-frame interactions during video encoding; it does not need to be the final conditioning memory for language generation. We therefore select temporal sinks using video-only attention scores and make no assumptions about how text queries might alter sink identities. 

We formalize the concentration and slow-evolution assumptions into testable mechanisms, and probe these claims empirically in Section~\ref{sec:sup_assumptions} of the supplementary material. That analysis, together with the main results in Section~\ref{sec:results}, provides evidence that the resulting importance-based approximation works well across tested parameter scales, cache sizes, and model families.

\subsection{Method Overview}

\paragraph{Preliminaries.}
Let $A \in \mathbb{R}^{L \times L}$ denote the attention matrix for a given layer and head, where $A_{i,j}$ is the normalized attention from query token $i$ to key token $j$, and let $V \in \mathbb{R}^{L \times d}$ be the corresponding value vectors. The full self-attention output for token $i$ is
{\small$$
y_i^{\text{full}} = \sum_{j=1}^{L} A_{i,j} V_j.
$$
}Video tokens are grouped into frames; for $N$ video frames with $T$ tokens per frame we have $L = TN$ video tokens.

\paragraph{Streaming cache construction with two memory states.}
Let a video consist of $N$ frames, each yielding $T$ visual tokens. We build the video representation in a streaming fashion, processing frames in order and constructing a key-value (KV) cache incrementally. The method maintains two per-layer KV states:
(i) a \emph{detailed state} (\texttt{dstate}) that stores all per-frame video tokens and is used for final text decoding; and
(ii) a small \emph{compressed state} (\texttt{cstate}) of fixed capacity that is used only as cross-frame context during cache construction.

Concretely, for each transformer layer $\ell \in \{1,\dots,L_{\text{layers}}\}$ we maintain
{\small\begin{equation}
  D_n^\ell = \{(K_{1:n}^\ell, V_{1:n}^\ell)\}, \quad
  C_n^\ell = \{(\bar{K}_n^\ell, \bar{V}_n^\ell)\}, \quad
  |\bar{K}_n^\ell| = |\bar{V}_n^\ell| = B.
\end{equation}
}where $D_n^\ell$ contains \emph{all} visual tokens accumulated up to frame $n$ (and thus grows as $O(nT)$), while $C_n^\ell$ is a compressed memory with a fixed budget $B$ (e.g., 1024 tokens). Importantly, $C_n^\ell$ is the \emph{only} information from frames $< n$ that frame $n$ can attend to during cache building. The detailed cache $D_n^\ell$ is updated by appending the current frame tokens after each step, and is never queried by future frames during cache building. By contrast, the compressed state is refreshed after every frame, so it evolves over time even though its capacity remains fixed.

\paragraph{Key insight: dynamic compressed state.}
Our method relies on the assumption that the tokens useful for processing frame $n$ are either (i) tokens that were already useful for frame $n\!-1$ (and therefore retained in $C_{n-1}^\ell$), or (ii) tokens from the current frame $n$.
This makes the compressed state \emph{dynamic} rather than fixed across the whole video: we are not attempting to summarize the entire past into a single static memory.
Instead, we maintain a small working set that is refreshed every frame so that frame $n$ has access to the information it needs. 

\subsection{Per-frame cache-builder forward pass}
Let $X_n \in \mathbb{R}^{T \times d}$ denote the hidden states of the visual tokens for frame $n$ at the input of a given layer. The cache builder computes, at each layer $\ell$, queries for the current frame and keys/values for both the current frame and the compressed memory:
{\small $$
Q_n^\ell = X_n W_Q^\ell, \qquad
K_n^\ell = X_n W_K^\ell, \qquad
V_n^\ell = X_n W_V^\ell, \qquad
\bar{K}_{n-1}^\ell, \bar{V}_{n-1}^\ell \in C_{n-1}^\ell.
$$}
We then perform attention for frame $n$ against the concatenation of the compressed memory and the current frame tokens:
{\small \begin{equation}
    \mathrm{Attn}\bigl(
    Q_n^\ell,
    [\bar{K}_{n-1}^\ell; K_n^\ell],
    [\bar{V}_{n-1}^\ell; V_n^\ell]
    \bigr)
    = \mathrm{softmax}\!\left(
    \frac{Q_n^\ell [\bar{K}_{n-1}^\ell; K_n^\ell]^\top}{\sqrt{d_h}} + M_n
    \right)\cdot [\bar{V}_{n-1}^\ell; V_n^\ell]
\end{equation}
}where $d_h$ is the head dimension and $M_n$ is the appropriate causal and modality mask (for cache building we only require causal structure within the stream order, with all memory tokens preceding the current frame tokens).

This yields updated hidden states $X_{n,\text{out}}^\ell$ for the current frame, which are passed to the next layer. After the final layer, we append the per-layer $(K_n^\ell, V_n^\ell)$ to the detailed state:
{\small$$
D_n^\ell \leftarrow D_{n-1}^\ell \cup \{(K_n^\ell, V_n^\ell)\}.
$$}
Thus, the detailed cache of stored tokens grows linearly as the model processes more frames, but we avoid quadratic cross-frame attention by restricting cross-frame interaction to the fixed-size compressed state.

\paragraph{Updating the compressed state via attention-driven selection}
Finally, after processing frame $n$, we update the compressed state $C_n^\ell$ by selecting a fixed number of tokens to carry forward. Let $\mathcal{U}_n^\ell$ denote the candidate pool for compression at layer $\ell$. In \method we use an incremental pool consistent with the ``slowly-evolving sinks'' assumption:
{\small$$
\mathcal{U}_n^\ell = \bar{S}_{n-1}^\ell \cup \{1,\dots,T\}_{\text{(frame }n\text{ tokens)}},
$$
}where $\bar{S}_{n-1}^\ell$ indexes the tokens currently stored in $C_{n-1}^\ell$. We assign each candidate token $j \in \mathcal{U}_n^\ell$ an importance score based on \emph{video-only attention}. Let $A_n^\ell$ denote the attention weights produced when processing frame $n$ at layer $\ell$ (aggregated over heads and queries within the frame). One simple instantiation is
{\small$$
s_{n,j}^\ell = \frac{1}{T} \sum_{i=1}^{T} A_{n,i,j}^\ell,
$$
}where $A_{n,i,j}^\ell$ is the normalized attention from query token $i$ in frame $n$ to candidate key token $j$ (which may refer either to a memory token or a token in frame $n$). We then keep the top-$B$ candidates:
{\small\begin{equation}
\begin{aligned}
\bar{S}_n^\ell &= \mathrm{TopK}\bigl(\{ s_{n,j}^\ell : j \in \mathcal{U}_n^\ell \},\, B\bigr),\\
C_n^\ell &\leftarrow \{(\bar{K}_n^\ell, \bar{V}_n^\ell)\}
= \{(K_j^\ell, V_j^\ell) : j \in \bar{S}_n^\ell\}.
\end{aligned}
\end{equation}
}This procedure realizes a fixed-capacity temporal cache that is refreshed each frame by evicting low-importance memory entries and admitting newly salient tokens from the current frame. In all experiments we compute scores from video-only attention statistics, per the final assumption.

\paragraph{Virtual sequence length and cache positions.}
Because $C_n^\ell$ is capacity-limited, the number of keys stored in the compressed state differs from the number of tokens seen so far in the stream. However, positional encodings (RoPE~\cite{su2024roformer}) depend on the logical position in the stream, not on the current cache size. We therefore maintain a \emph{virtual sequence length} $L_n$ that counts all tokens processed up to time $n$
independent of the number of tokens retained in $C_n^\ell$. When constructing attention for frame $n$, RoPE positions and cache positions are derived from $L_n$ rather than from the physical cache length of $C_n^\ell$.

\paragraph{Consistent RoPE scaling across cache building and generation.}
When the total sequence length exceeds the base model's trained maximum context, RoPE scaling must be applied \emph{consistently} to all keys/values that will be reused during generation. In particular, the video KV cache must be compatible with subsequently generated tokens. To ensure this, we:
(i) determine the maximum expected sequence length for the entire run (video frames plus prompt plus maximum generation length) prior to cache building;
(ii) activate the corresponding RoPE scaling configuration before the first frame is processed; and
(iii) keep the same scaling active for the full duration of cache building and text decoding.

Formally, let $\phi(\cdot; \alpha)$ denote the RoPE embedding function with scaling parameters $\alpha$ (e.g., YARN~\cite{peng2024yarn}). We apply
{\small$$
Q \leftarrow \phi(Q; \alpha),\quad K \leftarrow \phi(K; \alpha)
$$
}with the \emph{same} $\alpha$ for both cache building and generation. Changing $\alpha$ after building the KV cache would make the cached $K/V$ incompatible with newly rotated queries/keys and leads to severe degradation; thus $\alpha$ is fixed end-to-end for each evaluation run.

\paragraph{Decoding using the detailed cache.}
After the streaming cache builder finishes processing all frames, we run standard autoregressive decoding for the text output. Crucially, decoding uses the detailed state $D_N^\ell$ (all video tokens) as the conditioning KV cache. The compressed state $C_N^\ell$ is not used during decoding; it exists solely to approximate cross-frame interactions during cache construction. This design matches the assumption that the temporal cache only needs to be an approximation for video encoding, since the final generation stage conditions on the full set of stored video tokens.

\section{Results}
\label{sec:results}

\paragraph{Experimental Setup.}
Our setup isolates the effect of the self-attention approximation while keeping the rest of the inference stack fixed.
All experiments use Hugging Face~\cite{wolf2020transformers} model implementations and default prompts from the lmms-eval suite~\cite{zhang2024lmmseval}.
We evaluate on VideoMME~\cite{fu2024video} (subtitles-free setting), MLVU~\cite{zhou2024mlvu}, and  OVOBench~\cite{li2025ovobench} (\emph{Real-Time Visual Perception} subset).
Following common long-video evaluation protocol~\cite{eagle}, we sample video at 1 FPS and cap each example at 512 frames.
We first refactor execution into two stages: video prefill and language generation.
We then implement a streaming version of the full self-attention baseline so it can process very long sequences.
This conversion is mathematically exact and, with linear memory attention kernels (e.g., SDPA/FlashAttention~\cite{dao2023flashattention2}) keeps peak memory usage manageable, although per frame compute still grows linearly with processed context so late frames become very slow.
From that baseline, we modify only the attention operation to obtain \sliding and \method, so differences in performance can be attributed only to the approximation itself.
RoPE scaling, data loading, prompts, and generation hyperparameters are matched across methods. 
Unless stated otherwise, we report efficiency primarily in terms of measured FLOPs (profiled using PyTorch~\cite{paszke2019pytorch} profiler and verified against theoretical calculations) and asymptotic scaling, which are stable across hardware and kernel implementations.
Section~\ref{sec:sup_walltime} of the supplementary also includes a wall-time comparison showing that, even when Full SA uses FlashAttention-2~\cite{dao2023flashattention2} and \method uses an eager attention path during cache building, the constant per-frame cost of \method still overtakes the linearly increasing cost of full self-attention at sufficiently long sequence lengths.
We additionally provide a custom Triton~\cite{tillet2019triton} kernel in Subsection~\ref{sec:sup_triton} that replaces eager attention with a FlashAttention-2~\cite{dao2023flashattention2} based implementation 
that returns the per-key scores needed for token selection and moves the crossover to shorter sequences.

\begin{figure}[ht]
  \begin{center}
    \centerline{\includegraphics[width=\columnwidth]{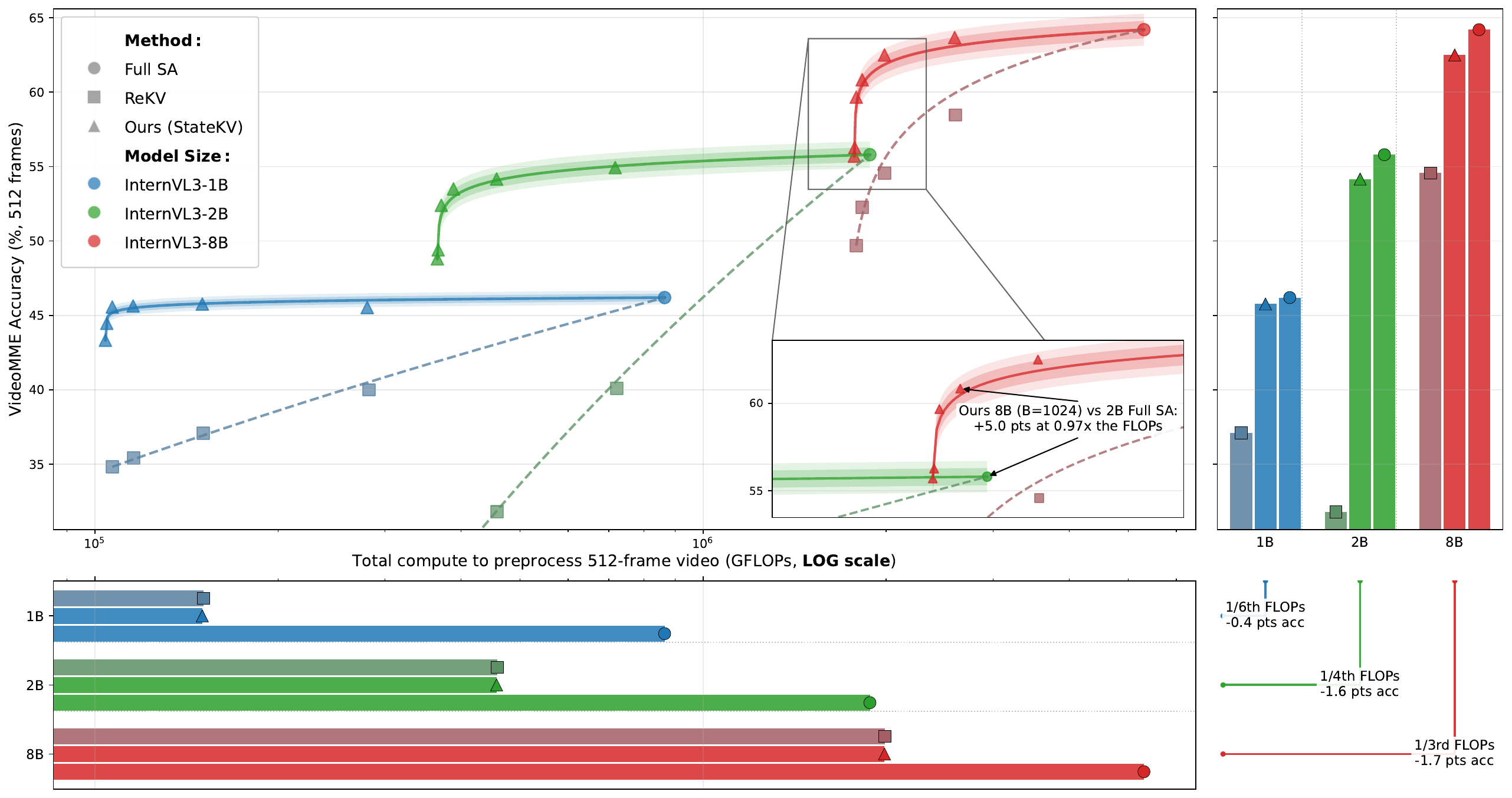}}
    \caption{Total compute to preprocess a 512-frame video (in GFLOPs) versus performance on VideoMME across three model sizes of the same model family (InternVL3 1B, 2B, 8B). Marker shape denotes which self-attention approximation (or Full SA) is used, while color denotes model size: circles are the 512-frame Full Self-Attention, triangles are \Method operating points at cache budgets $B \in \{16, 64, 256, 1024, 4096, 16384\}$, and squares are \sliding operating points at retained-frame budgets $R \in \{1, 4, 16, 64\}$. 
    Constant-FLOPs-per-frame methods reduce FLOPs compared to full self-attention while preserving accuracy. Of these, \Method achieves accuracy competitive with full self-attention and enables larger models under the same budget. For instance, \method-8B with $B=4096$ achieves 62.5\% accuracy at similar compute cost as Full SA-1B (46.2\%). Compared to existing sliding-window based methods like \sliding, \method achieves superior accuracy at comparable compression levels. The right and bottom supporting bars use one representative operating point per method: \method with $B=4096$ and \sliding with a 16-frame window (plus Full SA). These are compute-matched, so the bars isolate the quality difference at similar compute.}
    \vskip -0.3in
    \label{pareto}
  \end{center}
\end{figure}

\paragraph{Pareto frontier at fixed long-video length.}
Fig.~\ref{pareto} shows performance versus total prefill compute for a 512-frame video, with multiple compression settings for both \method and \sliding. Within each color, moving across triangles corresponds to increasing the \method cache budget $B$, while moving across squares corresponds to increasing the \sliding recency window $R$; this makes the within-model compute-accuracy scaling explicit. As expected, stronger compression reduces FLOPs but also weakens the approximation to full self-attention. Even under this tradeoff, \method remains consistently closer to Full SA than \sliding at comparable compute, and traces a stronger frontier across budgets. 
Notably, the \method operating points follow a smooth log-linear relationship between compute and accuracy, which enables predictable test-time scaling.

\paragraph{FLOPs reduction enables larger backbones.}
The practical implication of this frontier is that compute savings can be reinvested in model scale.
Although any approximation introduces some loss relative to exact Full SA for the same backbone, the loss for \method is small enough that moving to a larger model typically more than compensates for it. 
As a result, running larger models with \method yields substantially higher VideoMME accuracy at compute budgets comparable to or below smaller Full SA baselines.
In Fig.~\ref{pareto}, this appears as overlaps between a larger model's \method log-linear compute-accuracy curve, and the operating points for the unmodified Full SA baseline.
The marginal-cost variant of Fig.~\ref{pareto}, provided in Section~\ref{sec:sup_marginal} of the supplementary material, contains a double overlap, indicating that \method allows practitioners to run a model two scales larger at a similar per-frame-cost to the Full SA variant (InternVL3-8B with \method for a similar per-frame cost as the InternVL3-1B baseline).

\begin{table*}[tb]
\centering
\small
\setlength{\tabcolsep}{8pt}
\resizebox{\textwidth}{!}{
\begin{tabular}{llccc}
\toprule
\textbf{Model} & \textbf{Method} & \textbf{VideoMME} & \textbf{MLVU} & \textbf{OVOBench (Real-Time)} \\
\midrule
\multirow{3}{*}{InternVL3-1B}
  & Full SA              & \slurm{14086701}{46.19}
                         & \slurm{14101481}{47.05}
                         & \slurm{14125736}{55.79} \\
  & \sliding~(R=16)      & \slurm{14091595}{37.11}
                         & \slurm{14132035}{33.44}
                         & \slurm{14132111}{37.75} \\
  & \method~(B=4096)     & \slurm{14100168}{45.8}
                         & \slurm{14132034}{46.35}
                         & \slurm{14132110}{55.44} \\
\cmidrule[\heavyrulewidth]{1-5}
\multirow{3}{*}{InternVL3-2B}
  & Full SA              & \slurm{14488060}{55.8148}
                         & \slurm{14488063}{56.6104}
                         & \slurm{14488065}{60.2151} \\
  & \sliding~(R=16)      & \slurm{14339608}{31.7778}
                         & \slurm{14339611}{5.4931}
                         & \slurm{14339614}{33.3333} \\
  & \method~(B=4096)     & \slurm{14339607}{54.1481}
                         & \slurm{14339610}{57.3099}
                         & \slurm{14339613}{61.0514} \\
\midrule
\multirow{3}{*}{QWen3-VL-2B}
  & Full SA              & \slurm{14231300}{58.6667}
                         & \slurm{14231301}{58.8321}
                         & \slurm{14231302}{60.454} \\
  & \sliding~(R=16)      & \slurm{14210922}{49.4444}
                         & \slurm{14210925}{45.5993}
                         & \slurm{14210928}{46.7145} \\
  & \method~(B=4096)     & \slurm{14210921}{58.00}
                         & \slurm{14210924}{57.7243}
                         & \slurm{14210927}{60.9319} \\
\cmidrule[\heavyrulewidth]{1-5}
\multirow{3}{*}{QWen3-VL-4B}
  & Full SA              & \slurm{14231305}{66.5926}
                         & \slurm{14231306}{68.9755}
                         & \slurm{14231307}{64.7551} \\
  & \sliding~(R=16)      & \slurm{14200471}{52.6296}
                         & \slurm{14200474}{49.9078}
                         & \slurm{14200477}{49.2234} \\
  & \method~(B=4096)     & \slurm{14200470}{65.8889}
                         & \slurm{14200473}{67.7987}
                         & \slurm{14200476}{64.8746} \\
\cmidrule[\heavyrulewidth]{1-5}
\multirow{3}{*}{InternVL3-8B}
  & Full SA              & \slurm{14092115}{64.19}
                         & \slurm{14101467}{61.14}
                         & \slurm{14125735}{71.45} \\
  & \sliding~(R=16)      & \slurm{14071127}{54.56}
                         & \slurm{14125723}{31.11}
                         & \slurm{14125741}{56.03} \\
  & \method~(B=4096)     & \slurm{14003772}{62.52}
                         & \slurm{14125722}{62.85}
                         & \slurm{14125740}{70.25} \\
\midrule
\multirow{3}{*}{Eagle2.5-8B}
  & Full SA              & \slurm{14330582}{69.8148}
                         & \slurm{14330585}{73.8454}
                         & \slurm{14452673}{69.2951} \\
  & \sliding~(R=16)      & \slurm{14330581}{58.7037}
                         & \slurm{14330584}{55.014}
                         & \slurm{14459236}{55.4361} \\
  & \method~(B=4096)     & \slurm{14330580}{67.963}
                         & \slurm{14330583}{70.5173}
                         & \slurm{14459235}{68.6977} \\
\midrule
\multirow{3}{*}{QWen3-VL-8B}
  & Full SA              & \slurm{14178821}{70.52}
                         & \slurm{14178822}{75.82}
                         & \slurm{14178825}{67.86} \\
  & \sliding~(R=16)      & \slurm{14157894}{55.52}
                         & \slurm{14178380}{50.91}
                         & \slurm{14178353}{53.88} \\
  & \method~(B=4096)     & \slurm{14157893}{68.11}
                         & \slurm{14178379}{71.38}
                         & \slurm{14178352}{64.99} \\

\bottomrule
\end{tabular}
}
\caption{Cross-backbone comparison across three long-video benchmarks. For each backbone, we compare exact full self-attention, the recency-based streaming baseline \sliding with a 16-frame retrieval window, and \method with cache budget $B=4096$; these \sliding/\method settings are chosen to be compute-matched. Across model families and scales, \method stays close to Full SA while consistently outperforming \sliding on VideoMME, MLVU, and OVOBench, indicating that importance-based memory is a stronger approximation to full long-range attention than a strict sliding-window prior.}
\label{tab:combined_benchmarks}
\vskip -0.2in
\end{table*}

\begin{figure*}[tb]
  \vskip 0.2in
  \begin{center}
    \centerline{\includegraphics[width=\textwidth]{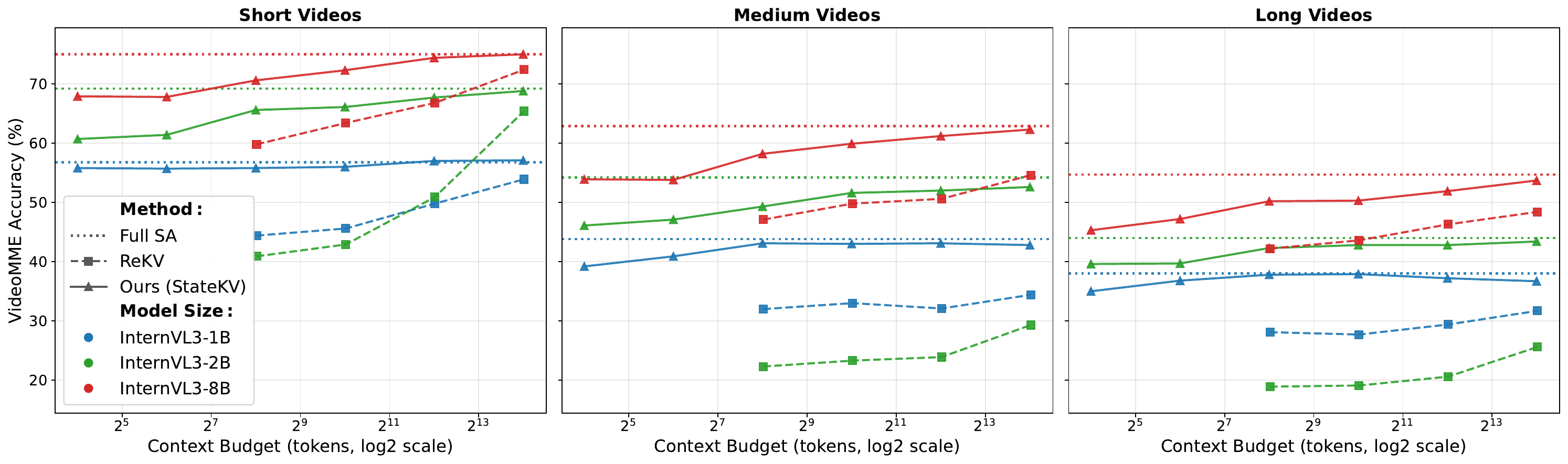}}
    \caption{Comparison of VideoMME accuracy across context budgets for InternVL3-1B/2B/8B. The dotted lines show Full SA (target behavior), while \sliding and \method trace budgeted approximations. Across short, medium, and long videos, \method stays consistently closer to the Full SA accuracy frontier than \sliding at comparable budgets, indicating a stronger approximation of full attention under constrained compute.}
    \label{fig:videomme_scaling_plot}
  \end{center}
  \vskip -0.3in
\end{figure*}

\paragraph{Cross-backbone ablation across families and scales.}
Table~\ref{tab:combined_benchmarks} compares full self-attention, \sliding, and \method across three video benchmarks and multiple backbones spanning different families and parameter scales. Averaged across these settings, \method stays close to full self-attention (within roughly a point on average) while consistently improving over \sliding by about 10 points on average. We observe the same trend on MLVU and OVOBench (Real-Time subset), indicating that the gain is not tied to a single backbone design or model size.
These cross-family results show that the same importance-based carried-memory intervention transfers across backbones, parameter counts, and cache budgets.
For a fair efficiency comparison, \sliding runs retrieving 16 frames and \method with cache budget $B=4096$ are compute-matched.

\paragraph{Scaling behavior.}
Fig.~\ref{fig:videomme_scaling_plot} studies accuracy as we increase the \method cache budget $B$ (or, for the baseline, the sliding-window retrieved frames). For all InternVL3 variants analyzed, \method improves monotonically across all video lengths and approaches, or in some settings matches, the full self-attention reference at the highest budgets (e.g., InternVL3-8B \method reaches 75.0\% on short videos, matching Full SA), demonstrating stable scaling with compute. Notably, the bigger models exhibit stronger scaling with cache size, with performance saturating at increasingly large budgets as model parameter counts increase. 
In contrast, sliding-window attention scales more slowly and remains below full attention at comparable budgets, with 5-10 point gaps across settings, suggesting that expanding only a local window does not recover the global-context benefits captured by \method. 
\Method achieves these gains while maintaining linear-in-frames video prefill computational cost, whereas Full SA remains costly on long videos.

\paragraph{Sliding-window instability across settings.}
Across our experiments, \sliding exhibits unstable behavior in multiple settings: InternVL3-2B shows systematic degradation across operating points in both Fig.~\ref{pareto} and Fig.~\ref{fig:videomme_scaling_plot}, occasionally underperforming even the 1B variant; InternVL3-8B shows similar instabilities on MLVU (Table~\ref{tab:combined_benchmarks}); and performance gaps relative to full attention remain substantial even at the highest sliding-window budgets tested. These observations suggest that strict recency-based approximations can be a poor match to the attention patterns learned by certain backbones.
We verified that this degradation is reproducible under the same shared implementation used across model scales and datasets. 
Subsection~\ref{sec:sup_recency} of the supplementary material discusses how \method's importance-based selection better approximates full self-attention patterns across these varied settings.

\begin{figure*}[tb]
  \vskip 0.2in
  \begin{center}
    \centerline{\includegraphics[width=\textwidth]{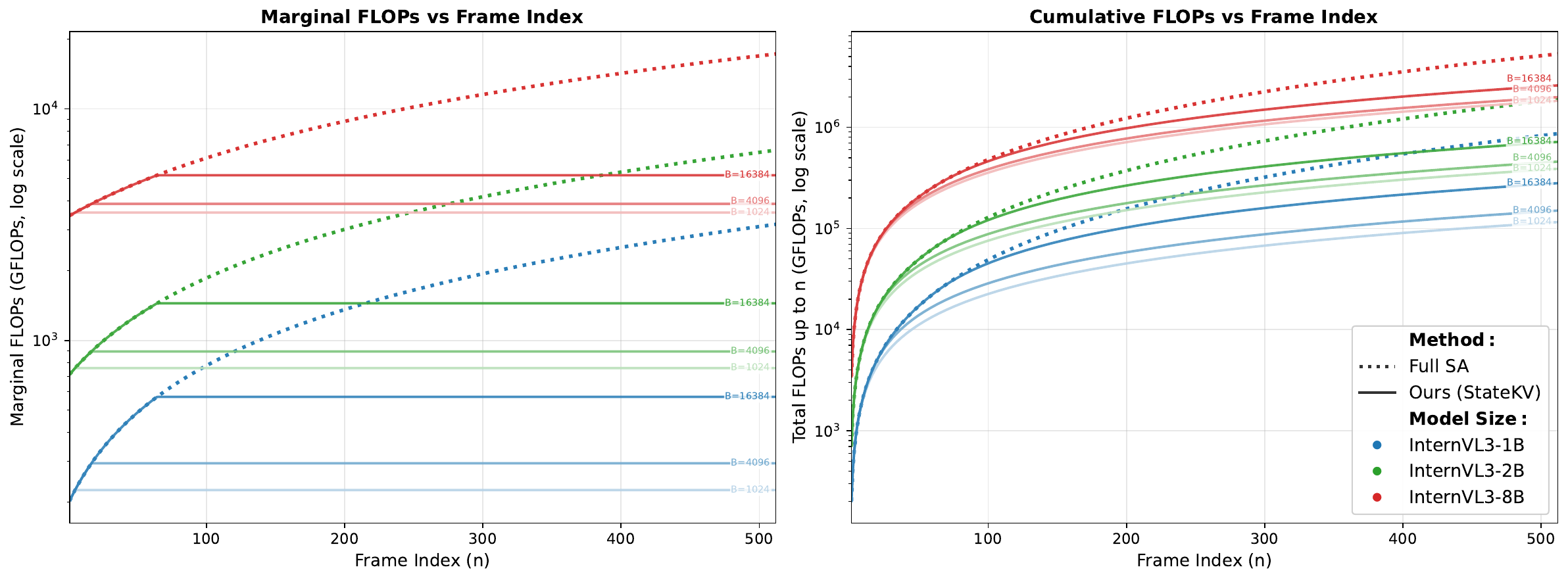}}
    \caption{Compute cost versus frame index. Left: marginal FLOPs per frame. Right: cumulative FLOPs. Dotted curves denote full self-attention and solid curves denote \method.}
    \label{flops}
  \end{center}
  \vskip -0.3in
\end{figure*}

\paragraph{Compute break-even behavior.}
Fig.~\ref{flops} highlights intersection points between dotted (Full SA) and solid (\method) curves, which mark compute break-even regimes across model sizes. In the left plot (marginal FLOPs per frame), each intersection gives the frame index beyond which adding one more frame is cheaper with a larger \method model than with a smaller Full SA baseline. In the right plot (cumulative FLOPs), intersections indicate the video length beyond which total compute is lower for the larger \method model over the full processed video. This distinction matters operationally: marginal cost is the relevant quantity for streaming-oriented settings, while cumulative cost is the relevant quantity for long-video processing. 
The key implication is that, because unmodified Full SA has quadratic video-prefill cost while \method is linear in the number of frames, a break-even intersection must exist at sufficiently long durations for each compared pair.
In other words, beyond a model- and setup-dependent horizon, it is compute-favorable to run a larger \method model rather than a smaller quadratic baseline. We further illustrate how this gap would widen at longer horizons by extrapolating compute curves to 3600 frames (1 FPS over 1 hour) in Subsection~\ref{sec:sup_longhorizion} of the supplementary material (Fig.~\ref{fig:supp_flops_3600_linear_log}).

\section{Conclusion}

We presented \method, a linear-time approximation to long-video self-attention for video VLMs. Building on recent streaming-prefill methods such as \sliding, our method separates video prefill into a streaming cache-construction stage and final text-decoding over the retained detailed video state. The key improvement is to frame streaming video prefill as approximating full self-attention using a small set of tokens carried between frames, rather than imposing a strict sliding-window prior. This design is motivated by mechanistic evidence on tested cases suggesting that useful inter-frame attention is often concentrated on a relatively small set of tokens and that this set evolves gradually enough to be updated from the previous state together with the current frame. Across multiple backbones, model scales, and long-video benchmarks, \method stays closer to full self-attention than sliding-window style baselines while reducing asymptotic video-prefill cost from quadratic to linear in the number of processed frames.

The main consequence in our measured FLOPs-based analysis is a better compute-accuracy tradeoff. Because \method preserves accuracy more effectively under compression, its FLOPs savings can be reinvested into larger backbones, yielding operating points where a larger \method model is both cheaper than and more accurate than a smaller full-attention baseline. Across model families, parameter scales, and cache sizes, we observe the same qualitative trend: carrying a small, importance-based set of tokens between frames is a stronger approximation than a strict recency bias. Our supplementary analyses further indicate that the analyzed attention patterns are not well described by a pure sliding-window view of the past, which helps explain why strict sliding-window approximations can be weak in our setting.

This work also has limitations. The mechanistic validation of the assumptions can only be performed on existing models and tested inputs, so it is neither general nor fundamental: although the assumptions are reasonable and supported on the models we analyze, untested backbones or future models may not exhibit the same behavior. 
Natural next steps include broader analysis across more backbones and video regimes.


\bibliographystyle{splncs04}
\bibliography{main}

\clearpage
\appendix
\section{Empirical validation of the assumptions}
\label{sec:sup_assumptions}

Our method is motivated by two empirical assumptions stated in Sec.~\ref{sec:assumptions}: (i) for a given frame $n$, most useful \emph{inter-frame} attention is concentrated on a relatively small set of historical tokens from frames $< n$; and (ii) the useful temporal state evolves slowly enough that the next state can be recovered from the previous state together with the current frame. We probe these assumptions directly using the unmodified full-attention baseline, and organize the analysis around the specific claim each assumption makes.

\paragraph{Scope.}
This analysis is intentionally mechanistic rather than exhaustive.
We formalize the assumptions from section and provide a testable mechanism we then use to probe InternVL3-1B/2B/8B on a dedicated attention-analysis set of 16 long videos from the VideoMME training split. The cache budgets are
\[
B \in \{1,4,16,64,256,1024,4096,16384\}.
\]
The analyzed video IDs are listed below:
\begin{quote}
\small
\texttt{1NYQf\_OXDqI}, \texttt{PXxscnWG8QA}, \texttt{0RxMZBLeqRI}, \texttt{oue5A-7Hpx4}, \texttt{yh-EHgkFci4}, \texttt{HTv4z899xgA}, \texttt{0kRsiSdDFYg}, \texttt{5WIdIs3A9Ok}, \texttt{Ry2dJuJ-9UE}, \texttt{KTjeh5QPL0o}, \texttt{p84O3JAp\_IM}, \texttt{sxrx7oCrb3A}, \texttt{1wzgMHrkrys}, \texttt{WQn-c\_4dVWs}, \texttt{WB4giHwiulE}, \texttt{rhDdA-7gEhs}.
\end{quote}
For each video we sample 128 frames approximately uniformly over the full video span using the InternVL3 video loader. We therefore view the present subsection as representative supporting evidence for the assumptions on tested cases rather than as a claim that all video VLMs exhibit identical attention structure. The broader empirical case for the intervention comes from the main-paper ablations across model families, parameter counts, and cache sizes.

\paragraph{Shared protocol.}
In all tests we run the unmodified full-attention baseline frame by frame, extract the video-only attention weights produced when encoding frame $n$, aggregate them over the queries in frame $n$, and then form the statistics below. Head treatment is important here. The implementation of \method maintains and prunes sinks separately within each layer and KV head, but the present analysis is intended to validate the higher-level layerwise assumptions from Sec.~\ref{sec:assumptions} rather than to reproduce the exact per-head update rule. We therefore first sum attention over the queries in frame $n$ within each head, and then sum those per-head scores to obtain a single layer-level score for each key token. Unless otherwise stated, reported curves average over analyzed videos, frames, and layers.

\subsection{Assumption 1: concentration of inter-frame attention}

\paragraph{Claim.}
Sec.~\ref{sec:assumptions} assumes that, for each frame $n$, most useful cross-frame attention is concentrated on a relatively small set of historical tokens from frames $< n$. This is the empirical basis for replacing unbounded cross-frame memory with a fixed-capacity set of temporal sinks.

\paragraph{Methodology.}
Let $A_n^\ell$ denote the attention weights produced when encoding frame $n$ at layer $\ell$, after aggregating over heads and over the queries belonging to frame $n$. As in Sec.~\ref{sec:assumptions} attention for frame $n$ is computed with the full key set available to that frame, including tokens from frame $n$ itself. To test the temporal-sink assumption, however, we only measure the portion of that attention assigned to keys from frames $< n$. Let $\mathcal{H}_n$ denote those historical keys, and let $s_{n,j}^\ell$ be the aggregated attention received by historical token $j \in \mathcal{H}_n$. We define the \emph{historical attention mass} for frame $n$ and layer $\ell$ as
\begin{equation}
M_{n,\mathrm{hist}}^\ell = \sum_{j \in \mathcal{H}_n} s_{n,j}^\ell.
\end{equation}
More explicitly, if $A_{n,h,i,j}^\ell$ is the attention from query token $i$ in frame $n$ to key token $j$ in head $h$ of layer $\ell$, then the layer-level score used throughout this subsection is
\begin{equation}
s_{n,j}^\ell = \sum_h \sum_{i=1}^{T} A_{n,h,i,j}^\ell.
\end{equation}
For a budget $B$, let $T_{n,B}^\ell \subseteq \mathcal{H}_n$ be the top-$B$ historical tokens ranked by $s_{n,j}^\ell$. The concentration plots report
\begin{equation}
C_{n,B}^\ell = \frac{\sum_{j \in T_{n,B}^\ell} s_{n,j}^\ell}{M_{n,\mathrm{hist}}^\ell},
\end{equation}
that is, the fraction of total historical attention mass captured by the top-$B$ historical tokens. The concentration curves plot the average of $C_{n,B}^\ell$ as a function of $B$. The frame-distance heatmaps summarize where this historical mass lands as a function of temporal distance from the current frame. If Assumption 1 is correct, then $C_{n,B}^\ell$ should rise rapidly with $B$ and approach saturation well before $B$ reaches the full historical context.

\begin{figure*}[!htbp]
  \centering
  \includegraphics[width=\textwidth]{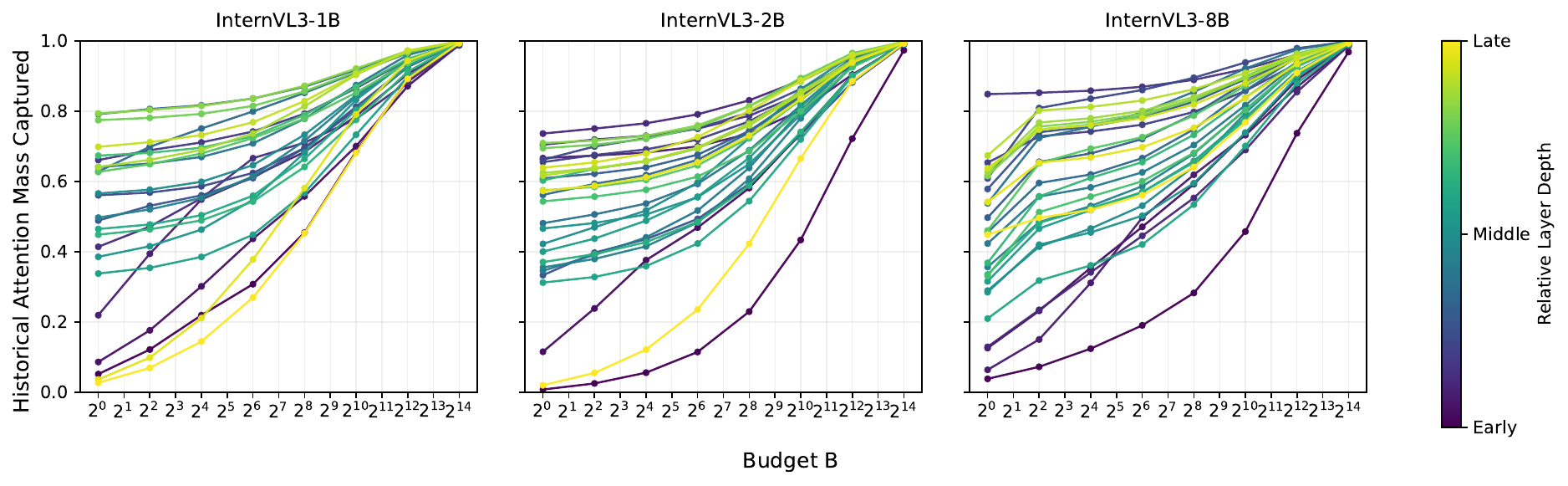}
  \caption{Validation of Assumption 1 on 16 long videos from the VideoMME training split, using 128 frames sampled approximately uniformly over each full video and budgets $B \in \{1,4,16,64,256,1024,4096,16384\}$. Each panel shows one model scale (InternVL3-1B/2B/8B). For each frame $n$, we compute attention with the full key set available to that frame, then report $C_{n,B}^\ell$, the fraction of total \emph{historical} attention mass assigned to the top-$B$ historical tokens from frames $< n$. Rapid saturation of these curves is the direct evidence for the bounded-sink assumption.}
  \label{fig:supp_assumption_concentration}
\end{figure*}

\paragraph{Current evidence.}
Assumption 1 is supported by these runs. As shown in Fig.~\ref{fig:supp_assumption_concentration}, the concentration curves rise monotonically with $B$ and saturate well before the full historical prefix is reached. In these 16-video, 128-frame long-video runs, the mean fraction of historical attention mass captured by the top-$B$ historical tokens is $0.71/0.69/0.72$ at $B=256$, $0.83/0.80/0.82$ at $B=1024$, and $0.93/0.93/0.93$ at $B=4096$ for 1B/2B/8B respectively. Thus, while the available historical context contains many more tokens than these budgets, a relatively small subset already captures most of the inter-frame attention mass.

The concentration plots also show a clear layerwise structure that persists across all three scales: historical attention concentration is high for most layers, while the first few and last few layers are the main outliers. This suggests that the bounded-sink behavior is strongest in the middle of the network and somewhat weaker near the ends.

\FloatBarrier

\subsection{Assumption 2: slow evolution of the temporal state}

\paragraph{Claim.}
Sec.~\ref{sec:assumptions} further assumes that the set of useful sinks evolves slowly enough that the useful state can be recovered from the combination of the previous state and the previous frame. This is the empirical basis for the incremental update rule of \method.

\paragraph{Methodology.}
To test the slow-evolution assumption, we distinguish this historical-only concentration analysis from the oracle state used for cache evolution. For a fixed layer $\ell$ and budget $B$, let $S_n^\ell$ denote the oracle top-$B$ state after processing frame $n$, now defined over \emph{all} seen tokens up to and including frame $n$. This is the oracle analogue of the compressed state described.
We then form the incremental candidate pool
$
S_n^\ell \cup \text{frame}_{n+1},
$
and measure how well it covers the next oracle state $S_{n+1}^\ell$. Writing
\begin{equation}
P_{n+1}^\ell = S_n^\ell \cup \text{frame}_{n+1},
\end{equation}
the \emph{candidate-pool recall} is
\begin{equation}
R_{n+1,B}^\ell = \frac{|S_{n+1}^\ell \cap P_{n+1}^\ell|}{|S_{n+1}^\ell|}.
\end{equation}
If $w_{n+1,j}^\ell$ denotes the attention mass assigned in frame $n+1$ to oracle token $j \in S_{n+1}^\ell$, then the \emph{weighted candidate-pool recall} is
\begin{equation}
\widetilde{R}_{n+1,B}^\ell
=
\frac{\sum_{j \in S_{n+1}^\ell \cap P_{n+1}^\ell} w_{n+1,j}^\ell}
{\sum_{j \in S_{n+1}^\ell} w_{n+1,j}^\ell}.
\end{equation}
This weighted version measures whether the highest-mass oracle tokens are preserved even if some lower-mass tokens are missed. We also report \emph{retention} and \emph{churn} between consecutive oracle states:
\begin{equation}
\mathrm{Retention}_{n+1,B}^\ell
=
\frac{|S_n^\ell \cap S_{n+1}^\ell|}{|S_n^\ell|},
\qquad
\mathrm{Churn}_{n+1,B}^\ell
=
\frac{|S_{n+1}^\ell \setminus S_n^\ell|}{|S_{n+1}^\ell|}.
\end{equation}
Here too the oracle states are formed from the same layer-level scores $s_{n,j}^\ell$, obtained by summing over queries and heads before ranking tokens. The candidate-pool recall plots report the average of $R_{n+1,B}^\ell$ or $\widetilde{R}_{n+1,B}^\ell$ as a function of $B$. The churn plot summarizes how the exact oracle membership changes with budget, while still showing the weighted recall to distinguish set turnover from loss of the highest-mass sinks.

\begin{figure*}[!htbp]
  \centering
  \includegraphics[width=\textwidth]{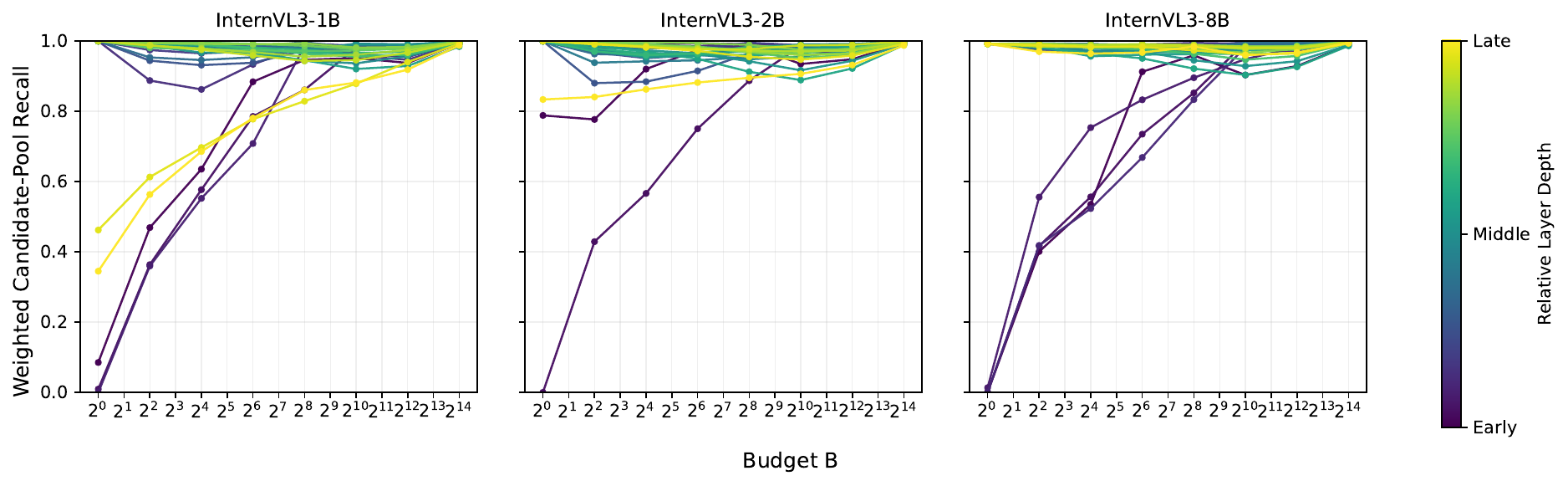}
  \caption{Primary validation of Assumption 2 on the same 16-video, 128-frame setting. Each panel shows one model scale (InternVL3-1B/2B/8B). For each layer and budget $B$, $S_n^\ell$ is the oracle top-$B$ state over all tokens seen up to and including frame $n$, and the plotted quantity is the weighted recall $\widetilde{R}_{n+1,B}^\ell$ of $S_{n+1}^\ell$ by the incremental candidate pool $S_n^\ell \cup \text{frame}_{n+1}$. High values mean that the most important members of the next oracle state are already present in the previous state plus the current frame, which is the specific slow-evolution claim used by \method.}
  \label{fig:supp_assumption_recall}
\end{figure*}

\paragraph{Current evidence.}
Assumption 2 is also supported, with the strongest evidence coming from the weighted metrics. As shown in Fig.~\ref{fig:supp_assumption_recall}, weighted candidate-pool recall is high across budgets, indicating that the candidate pool $S_n^\ell \cup \text{frame}_{n+1}$ usually contains the most important members of the next oracle state even when exact set overlap is not perfect. In these 1B/2B/8B runs, weighted candidate-pool recall is $0.90/0.95/0.92$ at $B=16$ and $0.96/0.97/0.96$ at $B=256$. In other words, the next oracle state is usually recoverable from the previous state together with the current frame, especially for the highest-mass sinks.

\begin{figure*}[!htbp]
  \centering
  \includegraphics[width=\textwidth]{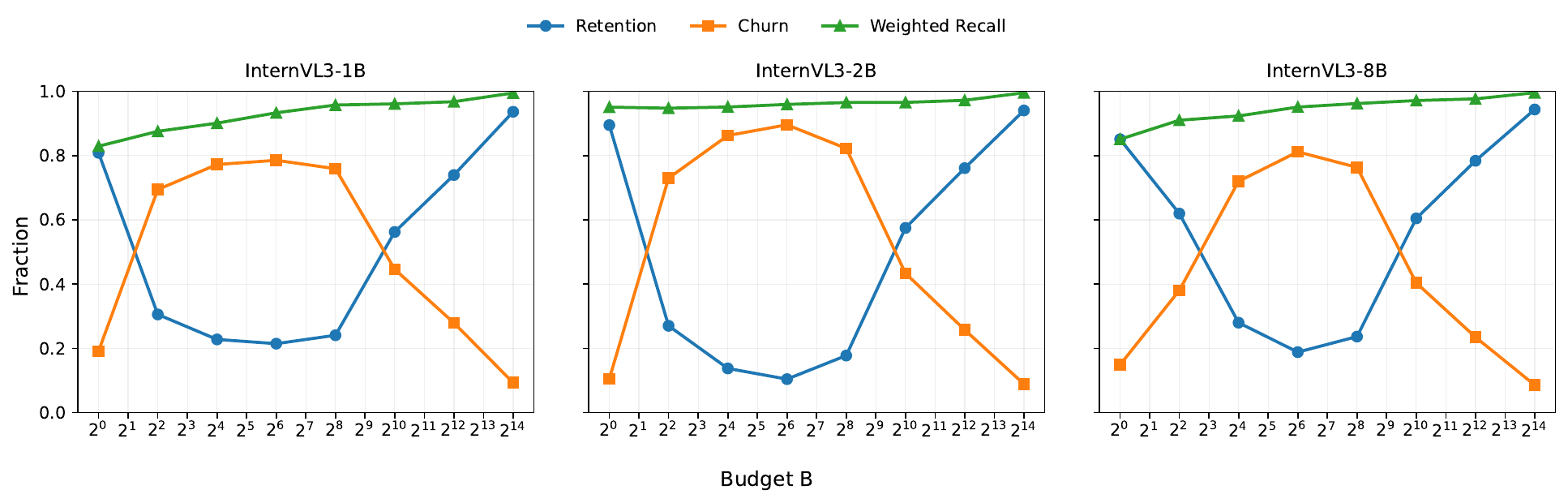}
  \caption{Additional analysis of Assumption 2 on the same 16-video, 128-frame setting. Retention measures the fraction of the oracle state that persists from frame $n$ to frame $n+1$, while churn measures the fraction of $S_{n+1}^\ell$ that is newly admitted. The repeated weighted-recall curve is included to distinguish exact set turnover from loss of high-mass sinks. The resulting pattern clarifies that the temporal state is not static, but changes in a structured way: a stable core persists even when a larger set of lower-mass sinks turns over.}
  \label{fig:supp_assumption_churn}
\end{figure*}

The retention and churn curves in Fig.~\ref{fig:supp_assumption_churn} refine this picture. Retention at $B=1$ is already $0.81/0.89/0.85$ for 1B/2B/8B, which means that the top-$1$ oracle token often persists from frame $n$ to frame $n+1$. At larger budgets, exact oracle membership changes substantially even while weighted recall remains high, indicating that the temporal state contains both a very stable core of high-importance tokens and a broader band of medium-importance tokens that changes more quickly. That decomposition is consistent with the incremental update rule in Sec.~3, which only needs the next useful state to be recoverable from the previous one and the current frame, not globally static over the whole video.

\FloatBarrier

\subsection{Supporting comparison: recency-based retention}
\label{sec:sup_recency}

\paragraph{Why this comparison is separate.}
The recency baseline is not itself one of the assumptions in Sec.~3, but it is the most relevant competing design family. This comparison therefore asks whether the attention patterns validated above would actually be useful for choosing which historical information to keep.

\paragraph{Methodology.}
We compare the attention-based concentration curve from Assumption 1 against a recency baseline that keeps the most recent $R \in \{1,4,16,64\}$ frames. Since each frame contributes 259 visual tokens in the current setup, these operating points correspond to token budgets of $259$, $1036$, $4144$, and $16576$. The attention-based curve reports the same quantity $C_{n,B}^\ell$ defined above, while the recency points report the fraction of historical attention mass captured when the retained historical set is forced to be a pure sliding window.

\begin{figure*}[!htbp]
  \centering
  \includegraphics[width=\textwidth]{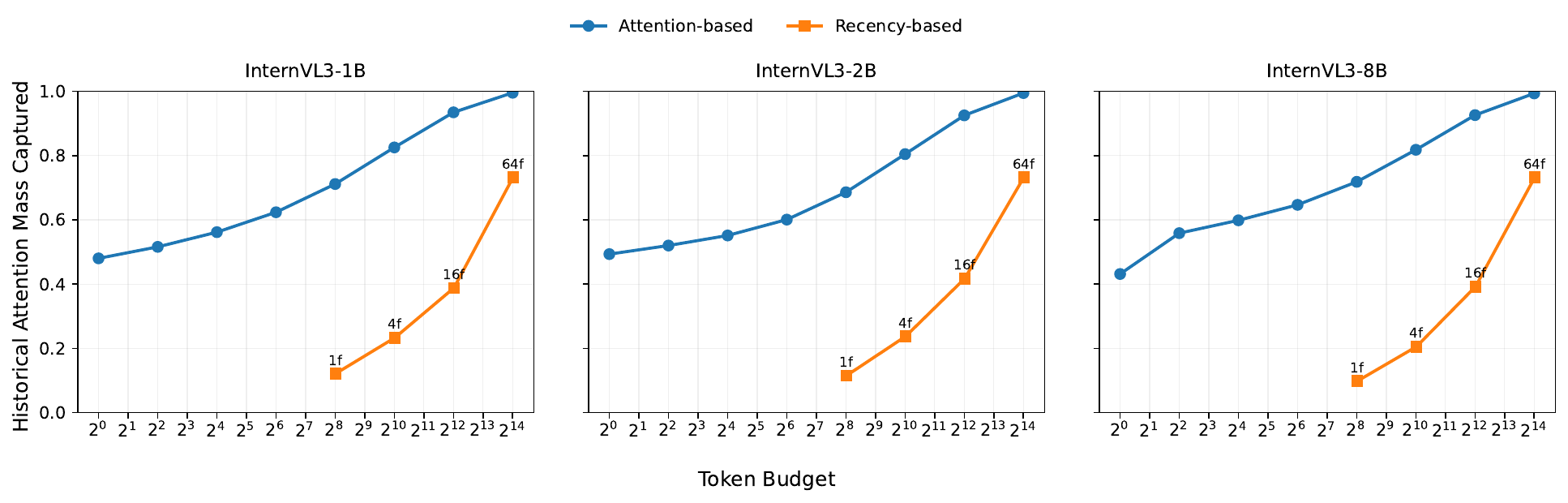}
  \caption{Supporting comparison to recency-based retention on the same 16-video, 128-frame setting. Each panel shows one model scale (InternVL3-1B/2B/8B). The attention-based curve reports $C_{n,B}^\ell$, the historical attention mass captured by top-$B$ historical tokens, while the recency baseline is evaluated at the explicit operating points corresponding to keeping the most recent $1$, $4$, $16$, or $64$ frames. This figure is not a direct assumption test, but it shows why the observed attention structure matters: if the main historical mass lies on a combination of recent frames and persistent long-range anchors, then a pure sliding window should underperform attention-based selection at comparable budgets.}
  \label{fig:supp_assumption_recency}
\end{figure*}

\paragraph{Current evidence.}
The direct comparison is shown in Fig.~\ref{fig:supp_assumption_recency}. We find that in these 1B/2B/8B runs attention-based selection captures substantially more historical attention mass than recency-based retention at comparable practical budgets: at $256$ tokens versus $1$ frame the gap is about $0.59/0.57/0.62$ for 1B/2B/8B, and at $1024$ tokens versus $4$ frames the gap is about $0.59/0.57/0.61$. 
This suggests that the dominant pattern is not pure recency.
Mechanistic analysis of the attention is consistent with that picture: the first and last available frames attract most of the historical attention mass for many of the attention heads.
This has a direct implication for sliding-window methods: they can preserve the near-past band, but they systematically discard the oldest-frame anchor.

\FloatBarrier

\section{Wall-time comparison under mismatched attention kernels}
\label{sec:sup_walltime}

Computing the temporal-sink scores used by \method requires access to per-layer attention weights (or sufficient statistics derived from them). In practice, this means the attention implementation must expose attention probabilities, which is typically not available in fused FlashAttention~\cite{dao2023flashattention2}/SDPA kernels. In our setting this is practical because cache building is performed one frame at a time: at each step we only compute attention between the current frame tokens and the compressed cache, rather than materializing attention over the full video-length sequence. We therefore use an attention path that can return attention weights during cache building, while keeping standard optimized attention for text decoding. This is a conservative comparison for \method: the full self-attention baseline can use FlashAttention-2, while \method is measured with eager attention during cache building.

All wall-time measurements in this subsection are collected on a single NVIDIA L40S GPU with batch size 1. For each point, we construct the preceding KV cache at the appropriate size for the method and frame index, time the model forward pass for processing one additional frame, discard warmup iterations, and then report the mean together with error bars from repeated measurements. Thus, the plotted quantity is the per-frame forward-pass latency for ``receive the preceding cache and process one more frame,'' rather than end-to-end dataset throughput.

\begin{figure*}[t]
  \centering
  \includegraphics[width=\textwidth]{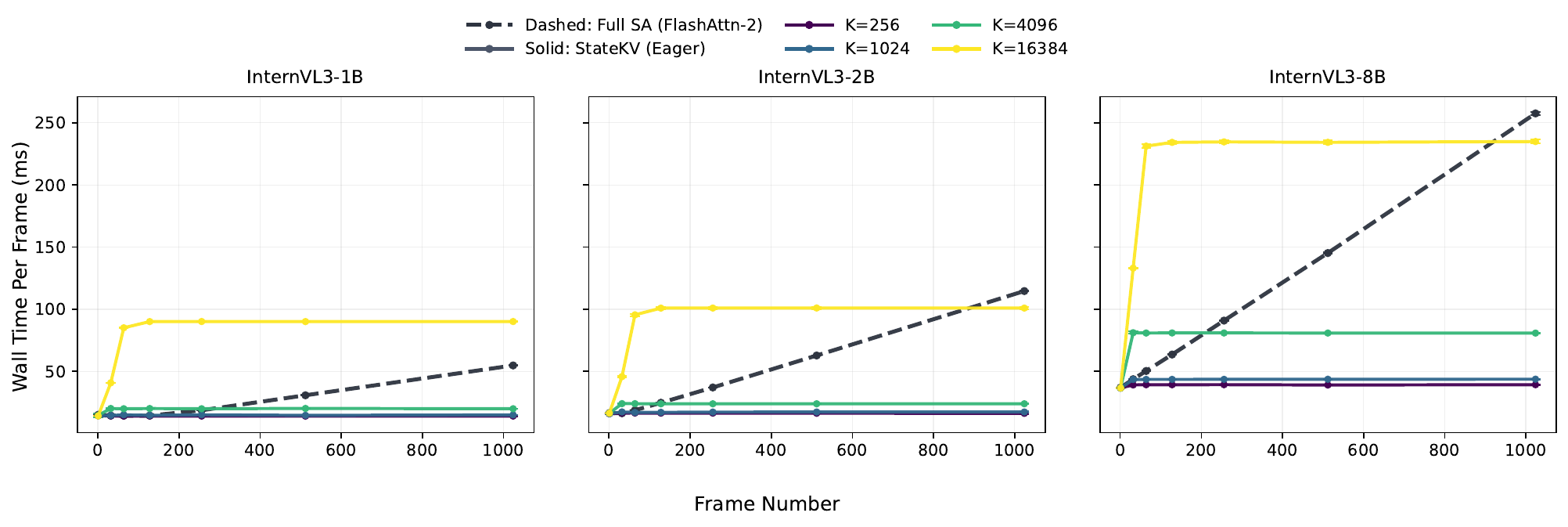}
  \caption{Measured wall time per frame versus frame index on a single NVIDIA L40S with batch size 1, comparing Full Self-Attention with FlashAttention-2 against \method with eager attention during cache building. For each point, we time the model forward pass for processing one additional frame given the preceding cache at the corresponding frame index, after warmup, and report standard-deviation error bars over repeated measurements. Solid colored curves sweep \method cache sizes $\{256, 1024, 4096, 16384\}$, while the dashed dark curve is the Full SA FlashAttention-2 baseline. Despite giving the baseline the more efficient attention kernel, the bounded per-frame cost of \method overtakes the linearly increasing cost of full self-attention for sufficiently long sequences.}
  \label{fig:supp_walltime_flash_vs_fastkv}
\end{figure*}

Figure~\ref{fig:supp_walltime_flash_vs_fastkv} shows that this kernel mismatch does not remove the asymptotic advantage of bounded per-frame cost. Even when full self-attention uses FlashAttention-2 and \method uses the less efficient eager attention path, the constant per-frame cost of \method eventually beats the linearly increasing per-frame cost of full self-attention. The crossover point depends on model size and compression budget, but the qualitative pattern is consistent across InternVL3-1B/2B/8B: once the sequence is long enough, fixed-budget memory dominates a kernel-efficient implementation whose cost still grows with context length.
This comparison is therefore conservative for \method, and future systems work should reduce its wall time further by moving the cache-building path to more optimized implementations, such as fused kernels or FlashAttention-style variants that expose the statistics needed for token selection.

\subsection{Triton kernel for fused attention score accumulation}
\label{sec:sup_triton}

The eager-attention path used above exposes per-layer attention weights so that \method can accumulate token-importance scores for cache pruning, but it materializes the full $Q \times K$ attention matrix in memory and cannot use FlashAttention-style tiling.
To close this gap we implement a custom Triton~\cite{tillet2019triton} kernel that performs attention in two passes without ever forming the full weight matrix.
The first pass is a standard tiled flash-forward: it computes the attention output $O$ and saves the per-query log-sum-exp (LSE) statistics.
The second pass re-reads $Q$, $K$, and the saved LSE to accumulate per-key score sums $s_k = \sum_q \exp(\text{score}_{qk} - \text{LSE}_q)$ directly into a compact $[\text{batch}, H_q, S_k]$ tensor, avoiding the $O(S_q S_k)$ allocation entirely.

\begin{figure*}[t]
  \centering
  \includegraphics[width=\textwidth]{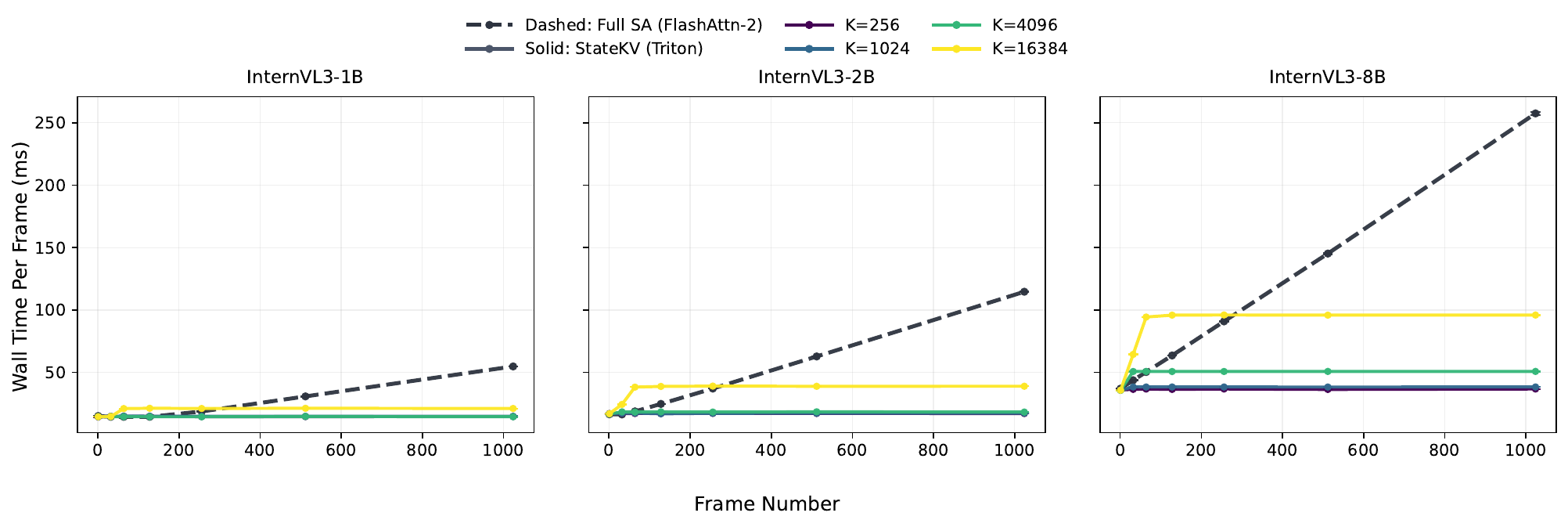}
  \caption{Measured wall time per frame versus frame index on a single NVIDIA L40S with batch size 1, comparing Full Self-Attention with FlashAttention-2 against \method with the Triton kernel during cache building. Format matches Fig.~\ref{fig:supp_walltime_flash_vs_fastkv}: solid colored curves sweep \method cache sizes $\{256, 1024, 4096, 16384\}$, while the dashed dark curve is the Full SA FlashAttention-2 baseline. Using the Triton kernel moves the crossover to shorter sequences compared to the eager-attention baseline.}
  \label{fig:supp_walltime_triton_vs_fastkv}
\end{figure*}

Figure~\ref{fig:supp_walltime_triton_vs_fastkv} shows the full-model level cost for eager versus Triton attention across model sizes and sequence lengths. 
Using the Triton kernel for cache building reduces \method's per-frame wall time across all model sizes and cache budgets, with the speedup growing at longer sequences where memory bandwidth for the full weight matrix dominates, and the crossover against full self-attention with FlashAttention-2 occurs at shorter sequences compared to the eager-attention baseline in Fig.~\ref{fig:supp_walltime_flash_vs_fastkv}.

\section{Additional Results}

\begin{figure*}[t]
  \centering
  \includegraphics[width=\textwidth]{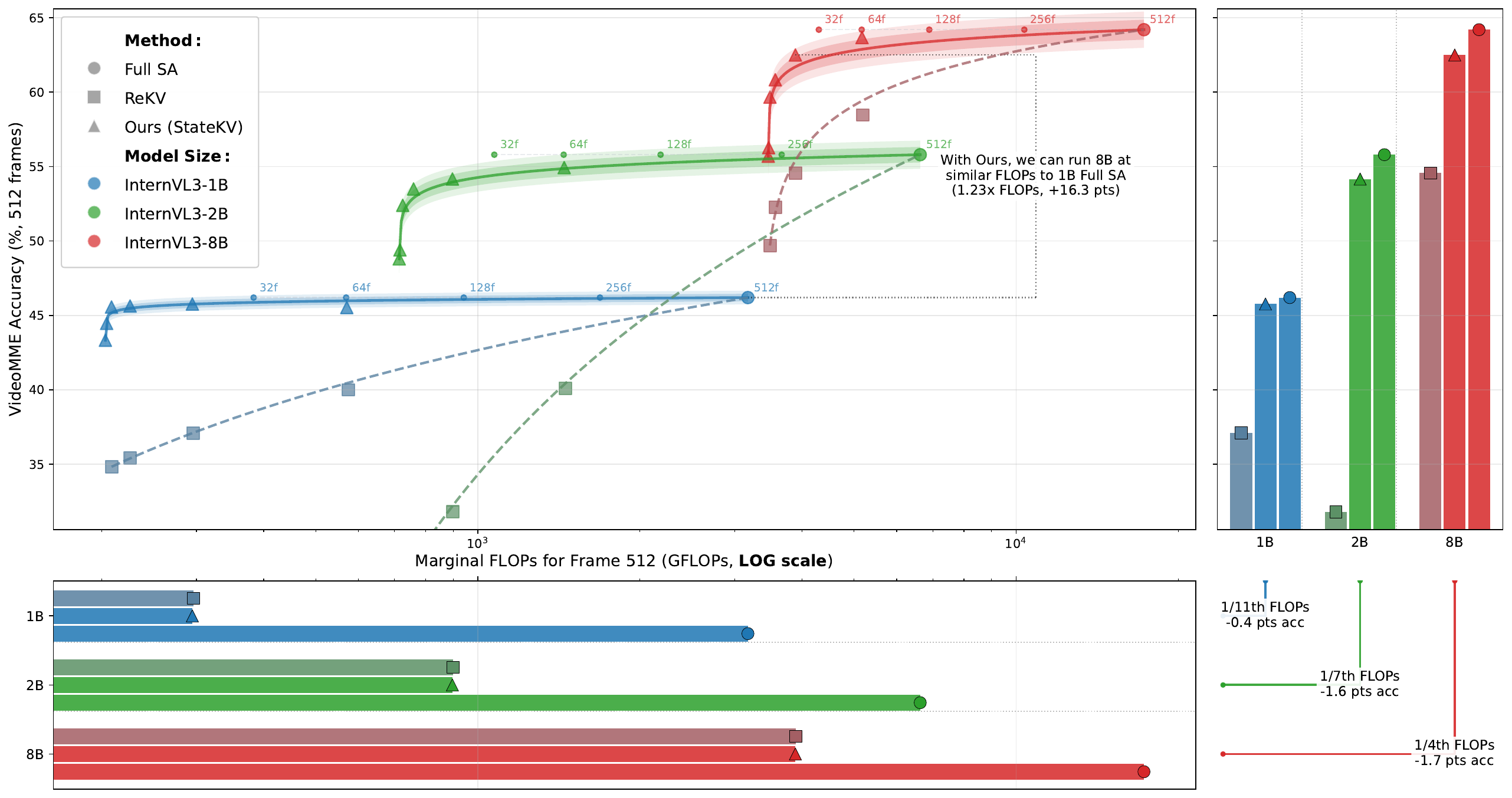}
  \caption{Marginal compute cost of processing another frame (in GFLOPs) versus performance on VideoMME across three model sizes of the same model family (InternVL3 1B, 2B, 8B). Marker shape denotes which self-attention approximation (or Full SA) is used, while color denotes model size: circles are Full Self-Attention measured at frames $32, 64, 128, 256,$ and $512$, triangles are \Method operating points at cache budgets $B \in \{16, 64, 256, 1024, 4096, 16384\}$, and squares are \sliding operating points at retained-frame budgets $R \in \{1, 4, 16, 64\}$. Full Self-Attention shows linear growth with video length, requiring increasingly more FLOPs as the prefix grows. In contrast, KV cache compression methods maintain constant marginal cost regardless of video length, trading off context preservation versus more aggressive compression for lower compute cost. \Method significantly reduces FLOPs compared to full self-attention at the same model size while maintaining competitive accuracy. Conversely, for a given compute budget \method allows us to run larger, more accurate models. For instance, \method-8B with $B=4096$ achieves 62.5\% accuracy at similar compute cost as Full SA-1B (46.2\%). Compared to existing sliding-window based methods like \sliding, \method achieves superior accuracy at all comparable compression levels. The right and bottom supporting panels summarize key per-experiment accuracy/FLOPs comparisons, and the lower-right panel reports the corresponding compute-accuracy tradeoff callouts.}
  \label{fig:supp_pareto_marginal_512}
\end{figure*}

\subsection{Marginal compute frontier}
\label{sec:sup_marginal}

Figure~\ref{fig:supp_pareto_marginal_512} is the marginal-cost companion to the accumulated-compute frontier in the main paper (Fig.~\ref{pareto}). The main figure asks how much total compute is required to preprocess a 512-frame video before generation, while the supplementary figure asks how expensive it is to process \emph{one more frame} once the preceding cache has already been built. This distinction matters for streaming-style deployment: accumulated cost captures the end-to-end prefill budget for a fixed-length video, whereas marginal cost captures the incremental cost paid as the sequence grows. As in the main figure, triangles trace increasing \method cache budgets $B$ within each model size and squares trace increasing \sliding window sizes, while the same color identifies the model backbone.

The qualitative conclusion is the same in both views. Full self-attention remains expensive because its per-frame cost grows with the number of previously seen tokens, while \method stays approximately constant at a fixed cache budget. The marginal view makes this especially explicit by collapsing each method to a per-frame operating cost and then comparing that cost directly against accuracy. In practice, this is the view most closely aligned with online or continuously growing video streams.

\begin{figure*}[t]
  \centering
  \includegraphics[width=\textwidth]{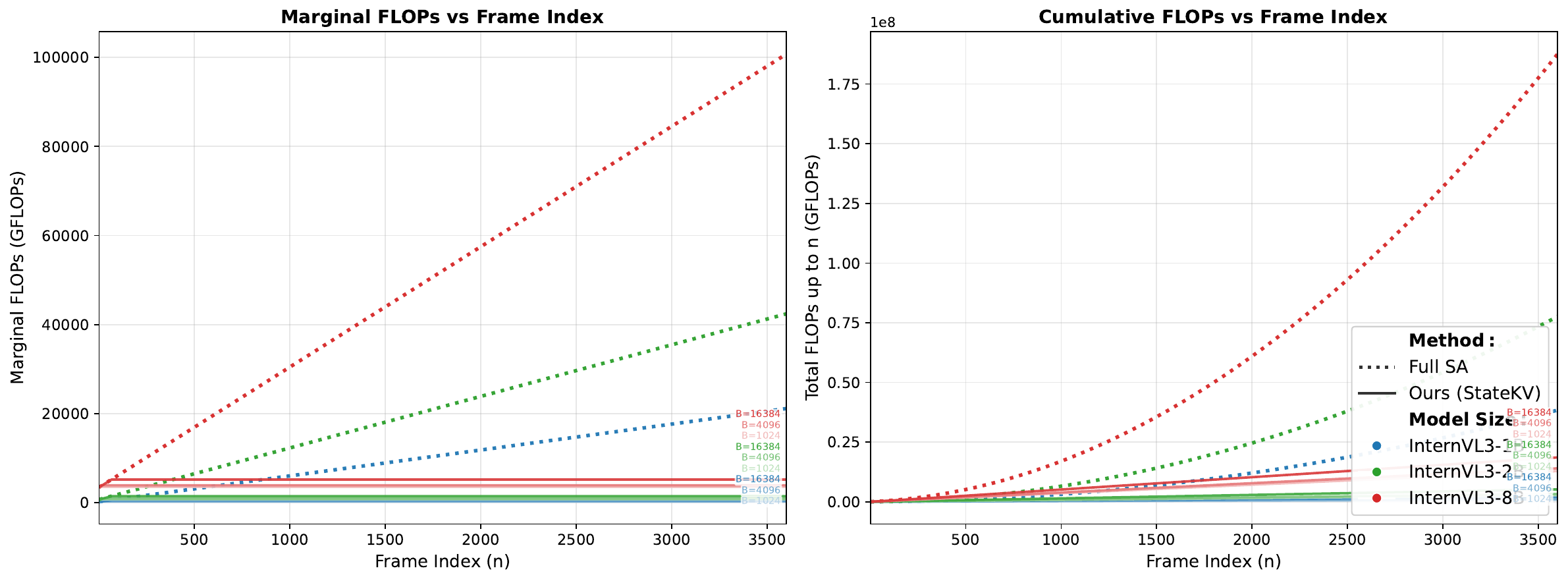}\\[0.6em]
  \includegraphics[width=\textwidth]{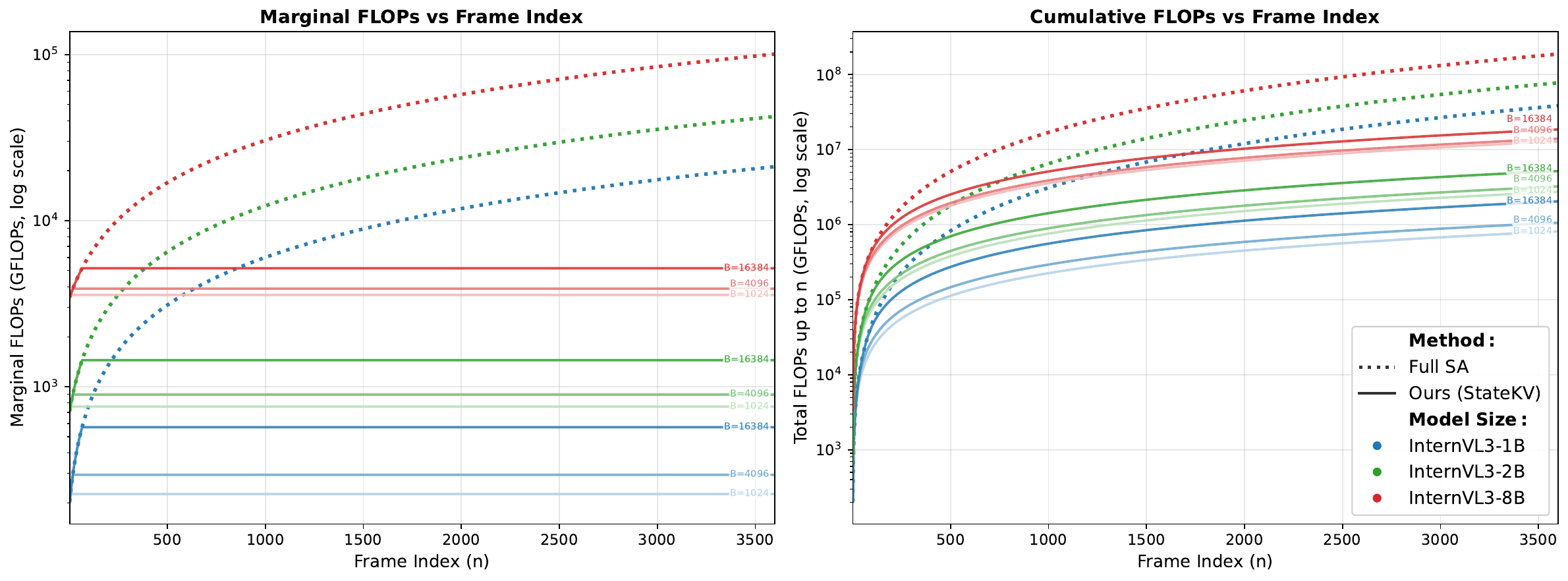}
  \caption{Compute cost versus frame index up to 3600 frames. Top: linear scale. Bottom: log scale. Left in each panel: marginal FLOPs per frame. Right: cumulative FLOPs. Dotted curves denote full self-attention and solid curves denote \method.}
  \label{fig:supp_flops_3600_linear_log}
\end{figure*}

\subsection{Long-horizon compute scaling}
\label{sec:sup_longhorizion}

Figure~\ref{fig:supp_flops_3600_linear_log} extends the compute-break-even analysis from the main paper (Fig.~\ref{flops}) to much longer horizons. The main figure already shows the intersection structure between larger \method models and smaller Full-SA baselines over the range most relevant to the benchmarked videos. The supplementary figure pushes that same analysis to 3600 frames, corresponding to roughly one hour of video at 1 FPS, to make the asymptotic separation easier to inspect.

Two points are worth emphasizing. First, the linear-scale panels are useful precisely because they make the asymptotic difference visually obvious: in the marginal-cost view, Full SA keeps getting more expensive as the prefix grows while \method stays approximately constant once the cache budget is fixed; in the cumulative-cost view, this becomes the familiar quadratic-versus-linear separation. The downside of this linear scale panel is that, at long horizons, the Full-SA curves grow so quickly that many break-even intersections become visually compressed because the smaller-\emph{y} region gets crowded. The log-scale panels complement this by making those intersections easier to inspect. In particular, they make clear that the crossover can occur even for very large compressed models, including the point where InternVL3-8B with the largest tested \method cache budget becomes cheaper than InternVL3-1B Full SA after around 1800 frames (approximately half an hour at 1 FPS). This is the strongest version of the scaling argument: for long enough videos, substantially larger backbones can become compute-favorable once the attention cost is linearized.

\end{document}